\pdfoutput=1

\documentclass[11pt]{article}

\usepackage{ACL2023}

\usepackage{times}
\usepackage{latexsym}

\usepackage[T1]{fontenc}

\usepackage[utf8]{inputenc}

\usepackage{microtype}

\usepackage{inconsolata}

\usepackage{caption}
\usepackage{graphicx}
\usepackage{subfigure}
\usepackage{threeparttable}
\usepackage{multicol}
\usepackage{multirow}
\usepackage{booktabs}
\usepackage{array}
\usepackage{amssymb}
\usepackage{amsmath}
\title{Logical Negation Augmenting and Debiasing for Prompt-based Methods}

\author{Yitian Li$^{1,2}$, Jidong Tian$^{1,2}$, \\ {\bf Hao He$^{1,2}$$^{\ddag}$ \and Yaohui Jin$^{1,2}$$^{\ddag}$} \\
        $^1$MoE Key Lab of Artificial Intelligence, AI Institute, Shanghai Jiao Tong University \\ $^2$State Key Lab of Advanced Optical Communication System and Network,\\ Shanghai Jiao Tong University\\
         {\tt \{yitian\_li, frank92, hehao, jinyh\}@sjtu.edu.cn}} 

\begin{document}
\maketitle
\begin{abstract}
Prompt-based methods have gained increasing attention on NLP and shown validity on many downstream tasks. Many works have focused on mining these methods' potential for knowledge extraction, but few explore their ability to make logical reasoning. In this work, we focus on the effectiveness of the prompt-based methods on first-order logical reasoning and find that the bottleneck lies in logical negation. Based on our analysis, logical negation tends to result in spurious correlations to negative answers, while propositions without logical negation correlate to positive answers. To solve the problem, we propose a simple but effective method, Negation Augmenting and Negation Debiasing (NAND), which introduces negative propositions to prompt-based methods without updating parameters. Specifically, these negative propositions can counteract spurious correlations by providing "not" for all instances so that models cannot make decisions only by whether expressions contain a logical negation. Experiments on three datasets show that NAND not only solves the problem of calibrating logical negation but also significantly enhances prompt-based methods of logical reasoning without model retraining.
\end{abstract}

\section{Introduction}

\begin{figure}[!tb]
\centering
\includegraphics[width=7.3cm]{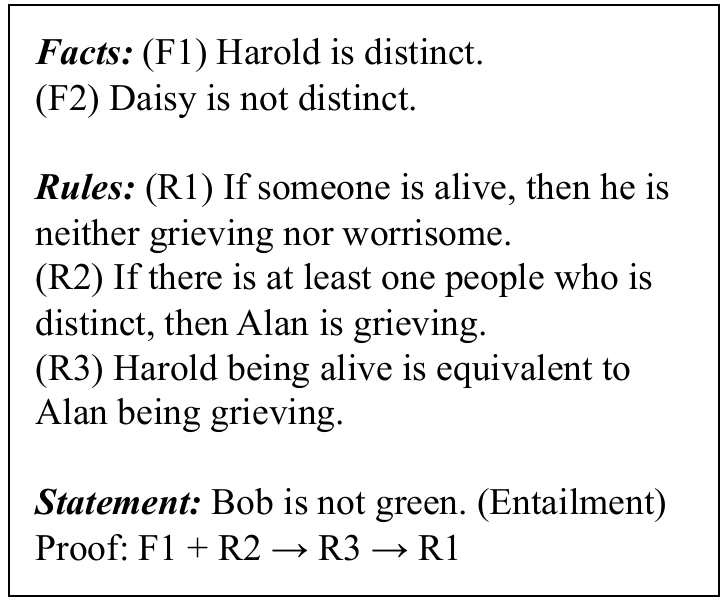}
\caption{An example from LogicNLI. Both RuleTaker and ProofWriter have similar forms.}
\label{fig1}
\end{figure}

Prompt-based methods~\cite{DBLP:journals/corr/abs-2108-13161,DBLP:conf/emnlp/PetroniRRLBWM19,DBLP:journals/tacl/JiangXAN20}, using pretrained language models (PLMs)~\cite{DBLP:conf/naacl/DevlinCLT19, DBLP:journals/corr/abs-1907-11692, DBLP:conf/nips/BrownMRSKDNSSAA20} and human-designed prompt templates for specific tasks, have achieved great success in knowledge-based natural language understanding (NLU)~\cite{DBLP:conf/acl/0010LLWWBCH22, DBLP:conf/acl/WangL020,DBLP:conf/blackboxnlp/YenicelikSK20,DBLP:conf/acl/JawaharSS19}. However, whether these methods can deal with logical reasoning in NLU lacks attention. In this work, we investigate whether prompt-based models can query logic processes from PLMs. To avoid the interference of extra knowledge, we perform analysis on three first-order logical NLI datasets without real-world knowledge: RuleTaker~\cite{DBLP:conf/ijcai/ClarkTR20}, ProofWriter~\cite{DBLP:conf/acl/TafjordDC21}, and LogicNLI~\cite{DBLP:conf/emnlp/TianLCX0J21} (an example is shown in Figure~\ref{fig1}). We find that prompt-based methods can leverage most logic forms in PLMs for downstream tasks except logical negation ($\neg$), as many studies have mentioned~\cite{DBLP:conf/naacl/HosseiniRBHSC21,DBLP:conf/acl/HossainHPB20}. 

Furthermore, we explore why prompt-based methods cannot deal with logical negation. Actually, logical negation is an operation on the true values of propositions instead of an indicator of a negative result~\cite{1991Classical}. For example, \textit{"I do not hate the movie"} does not mean negative sentiment about the movie. Instead, "not" here represents a negative operation on the negative verb "hate", so the sentiment about the example is positive. However, like other deep neural networks~\cite{DBLP:conf/naacl/HosseiniRBHSC21, DBLP:conf/acl/HossainHPB20}, prompt-based methods cannot process this operational feature effectively but tend to build strong spurious correlations between logical negation and negative labels. Specifically, without considering logic, statements with "not" are more likely to be classified as "Contradiction". On the other hand, statements without "not" are also incorrectly correlated to positive labels, including "Entailment". This phenomenon is named logically negative bias, which hinders machines from making correct logical reasoning.

To solve the problem, we propose \textbf{N}egation \textbf{A}ugmenting and \textbf{N}egation \textbf{D}ebiasing (NAND), which is a method that does not need to update parameters. NAND includes an augmentation module that takes advantage of negative propositions to compensate for instances without logical negation in statements. This strategy is simple but effective in alleviating the influences of the bias introduced by logical negation as all predictions take logically negative propositions into consideration. In addition, aiming at the open-world assumption (OWA) condition, we apply an empirical debiasing factor to balance the additional label, "Neutral". Finally, we test NAND on the three datasets mentioned above. Results exhibit that this method can eliminate the logically negative bias in logical reasoning without parameters updating, which matches prompt-based methods well.

Our main contributions include:
\begin{itemize}
    \item We analyze the effectiveness of prompt-based methods in logical reasoning and find that the bottleneck lies in logical negation.
    \item We further explore the reason for this phenomenon and find that logical negation tends to be incorrectly correlated to negative labels, while propositions without logical negation are more likely to correlate to positive ones. This phenomenon is named logically negative bias.
    \item We finally propose a simple but effective method, NAND, to remove the logically negative bias without parameter updating. We demonstrate the effectiveness of our method on three datasets. Our approach significantly boosts few-shot prompting baselines' performance, closes the gap with supervised models, and exhibits greater generalization.
\end{itemize}

\section{Related Work}
\subsection{Prompt}
Pre-trained language model prompting is popularized by recent work and has been demonstrated to be effective~\cite{DBLP:conf/emnlp/PetroniRRLBWM19,DBLP:conf/acl/CuiWLYZ21, DBLP:conf/chi/ReynoldsM21}. Prompting reformulates each task to match the pretraining objective to stimulate the rich knowledge hidden in the PLMs. For example, \citet{DBLP:conf/emnlp/PetroniRRLBWM19} use cloze-style to probe the commonsense knowledge that PLMs acquire during pretraining~\cite{DBLP:journals/corr/abs-2107-13586}. \citet{DBLP:journals/tacl/Ettinger20} assesses linguistic capacities by asking targeted questions. More complex abilities are also explored, such as symbolic reasoning~\cite{DBLP:journals/tacl/TalmorEGB20} and rare word understanding~\cite{DBLP:conf/aaai/SchickS20}. These mined knowledge and abilities are also used as an enhancement for NLP tasks~\cite{DBLP:conf/emnlp/GhosalMMP21,DBLP:conf/emnlp/ChakrabartyTM21}. Different manual prompt formats would result in differentiated accuracy~\cite{DBLP:conf/acl/LuBM0S22}. Therefore, \citet{DBLP:conf/emnlp/ShinRLWS20} have attempted to search for discrete prompt tokens automatically. With the continuous development, some work reveals that prompt-based methods also exploit superficial cues and lead to the bias of language models toward predicting certain answers~\cite{DBLP:conf/acl/KavumbaTO22,DBLP:conf/icml/ZhaoWFK021} using the same prompt formats. \citet{DBLP:journals/corr/abs-2105-11259} have applied logic rules by the sub-prompts to help relation classification.

\subsection{Negation}
Negation is a core construction in both linguistics and logic~\cite{DBLP:conf/naacl/HosseiniRBHSC21}. Despite being very successful, PLMs cannot always handle negation properly~\cite{DBLP:conf/acl/VasilakesZMA22}. In linguistics, negation is a phenomenon of semantic opposition~. \citet{DBLP:conf/acl/KassnerS20} found that these models fail at understanding negation through analyzing negated factual statements. Some work also shows that some negated triggers are biased towards contradiction~\citet{DBLP:conf/naacl/GururanganSLSBS18,DBLP:conf/emnlp/WallaceFKGS19,DBLP:conf/lrec/KhandelwalS20}. On the other hand, negation is first a phenomenon of semantic opposition~\cite{sep-negation}. ~\citet{DBLP:conf/acl/HossainHPB20} have proved that negation logic is difficult to be understood by neural networks.  \citet{DBLP:conf/emnlp/TianLCX0J21} also find the PLMs’ performance on negation logic is significantly worse than humans’. 

\begin{figure}[!tb]
\centering
\includegraphics[width=7.6cm]{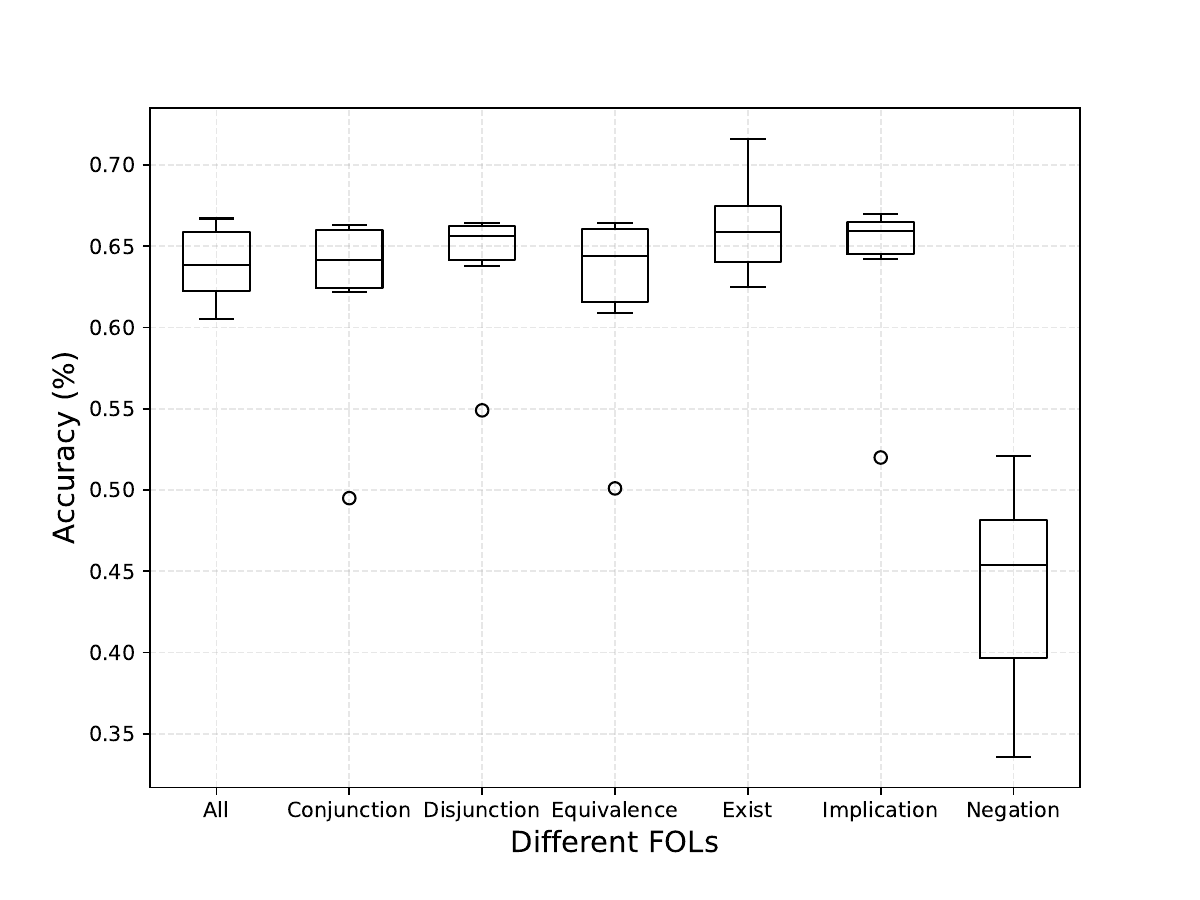}
\caption{Performance on each FOLs (\%). Here, we select eight different sets of prompt formats. For each formats, we plot BERT’s accuracy on differet FOLs from LogicNLI. Dots indicate abnormal results.}
\label{fig2}
\end{figure}

\section{Background}

\subsection{First-order Logical Reasoning}
In this work, we focus on first-order logic (FOL), one of the most widely used reasoning forms in NLU~\cite{DBLP:conf/iclr/YuJDF20,DBLP:journals/jair/Davis17}. It is a simple paradigm consisting of seven basic logics and there combinations (conjunctive~$\wedge$, disjunctive~$\vee$, negation~$\neg$, implication~$\rightarrow$, equation~$\equiv$, universal quantifier~$\forall$, and existential quantifier~$\exists$)~\cite{DBLP:conf/emnlp/TianLCX0J21}. We name $\neg$ logical negation to distinguish it from other types of negation.

Considering FOL reasoning in question answering systems, there are two world assumptions~\cite{REITER1981119} that result in different objectives. One is the \textbf{closed world assumption} (CWA), which is the presumption that what is not currently known to be entailment is contradiction. The other is the \textbf{open world assumption} (OWA), whose objective should distinguish false propositions from uncertain ones. Due to differences in world assumptions, our analysis and solutions are also different.

\subsection{Prompt-based Method}
Prompt-based methods include prompts and PLMs to complete different NLP tasks uniformly~\cite{DBLP:journals/corr/abs-2107-13586}. Prompts consists of placeholders for the training and test examples and a natural language description of the task~\cite{DBLP:conf/icml/ZhaoWFK021}, which are used to formulate tasks and activate PLMs to achieve predictions. For example, we use \textit{"Facts. Rules? [MASK], Statement."}, \textit{"Facts. Rules. [SEP] [MASK], Statement."}, etc. However, these methods suffer from fluctuation on different manual prompts~\cite{DBLP:conf/acl/GaoFC20}.

\section{Analysis of Prompt-based Methods on Logical Reasoning}

We perform an analysis of prompt-based methods' ability to mine PLMs' logical reasoning ability in this section. Specifically, we adopt different prompts on RuleTaker and LogicNLI and analyze their performance and variance. All of the prompts used in this study were chosen from the work of previous prompts on NLI. We observe that the bottleneck of prompt-based methods is logical negation. Furthermore, we explore why logical negation cannot always be understood and conclude logically negative bias that logical negation is likely to be correlated to negative labels.

\textbf{Prompt-based methods have certain capabilities to probe FOLs.} We use multiple prompt templates and two PLMs (BERT and RoBERTa), and choose the one of outstanding performance, whose results are shown in Table~\ref{Results1}. Firstly, prompt-based methods on all depths perform better than random guesses(50.0\%). Secondly, although the performance of prompt-based methods is inferior to that of fine-tuning methods, the gap between the two in out-of-domain generalization becomes smaller as the complexity increases. Considering depth-5, the advantage of fine-tuning methods over prompt-based methods is less than 4 points, but the latter does not require any extra training. Thirdly, we also experiment on AutoPrompt and find that its performance on logical reasoning is not satisfactory. These results show that prompt-based methods own the logical reasoning ability to some degree but cannot fully make logical reasoning in NLU.

\begin{figure}[t]
\centering
\includegraphics[width=7.5cm]{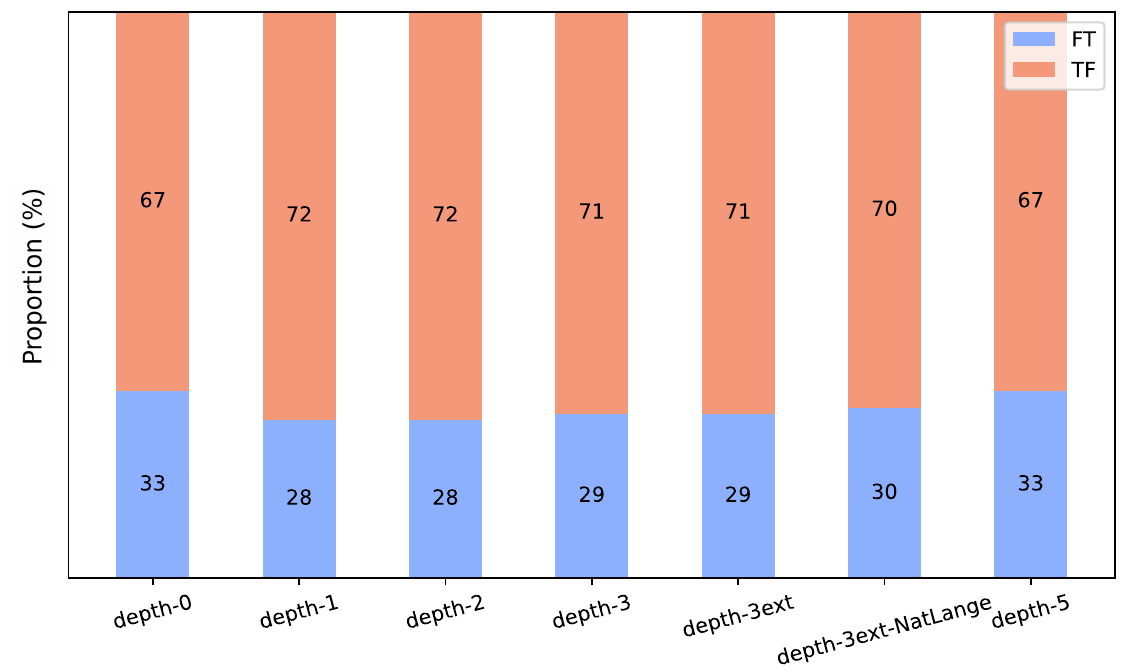}
\caption{Error statistics of the prompt-based method on the RuleTaker (RoBERTa). The results represent the proportion of different cases in which the model is misjudged. TF denotes that the model incorrectly interprets the "Entailment" label as "Contradiction", whereas FT denotes that the model incorrectly interprets the "Contradiction" label as "Entailment". It demonstrates that the findings are skewed toward specific responses.}
\label{figstat}
\end{figure}

\begin{figure}[t]
\centering
\includegraphics[width=7.5cm]{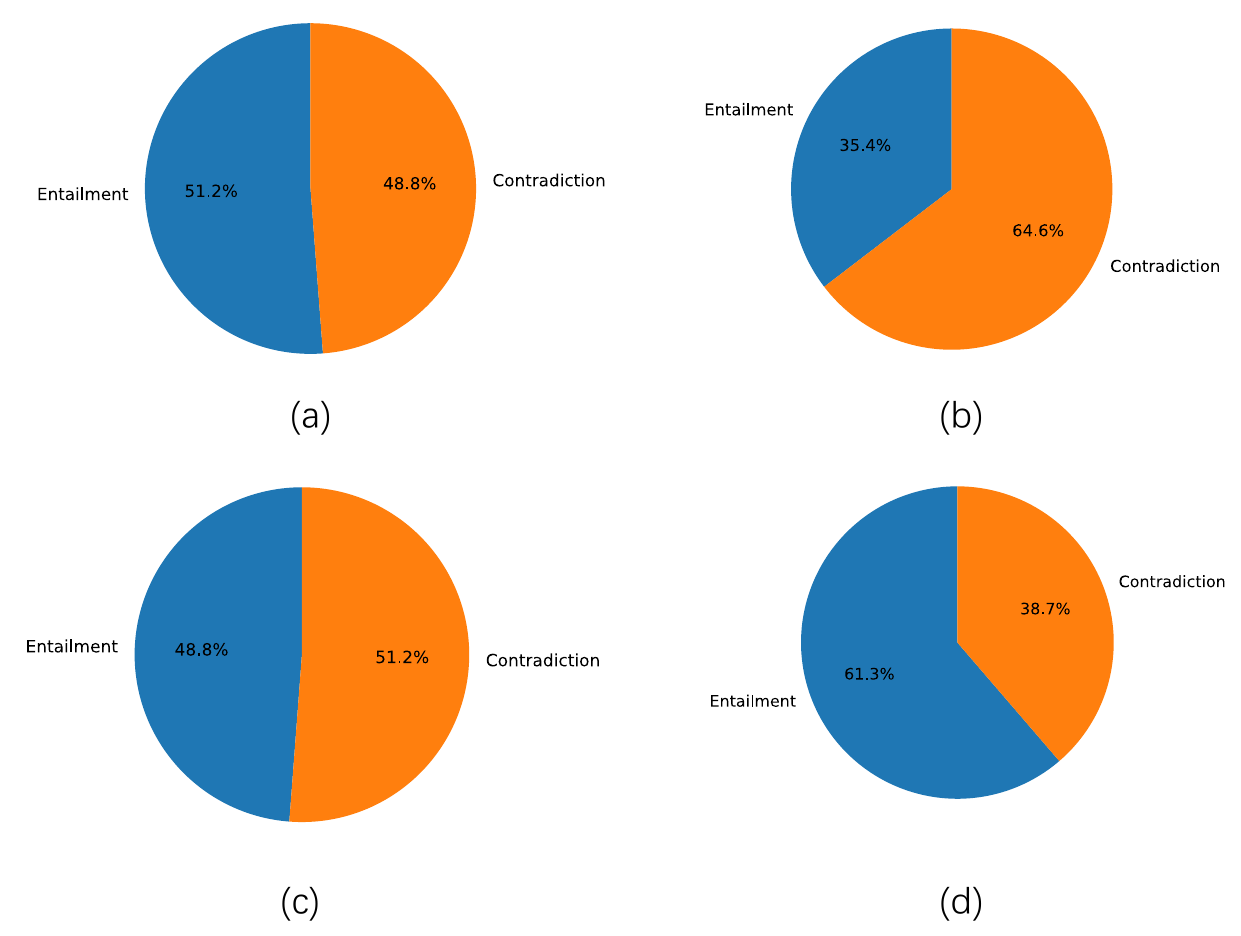}
\caption{Figure (a) shows the ratio of "Entailment" and "Contradiction" samples of statements with the negative word "not" in depth-0. Figure (c) shows the ratio of "Entailment" and "Contradiction" samples of statements without the negative word "not". "Entailment" and "Contradiction" represent the  labels of the data. Figure (b) shows the predictions of the prompt-based method on statements with the negative word "not" in depth-0 (RoBERTa). Figure (d) shows the predictions of the prompt-based method on statements without the negative word "not" in depth-0 (RoBERTa). "Entailment" and "Contradiction" represent the model's predictions. }
\label{figtj}
\end{figure}

\textbf{The bottleneck of prompt-based methods on FOLs lies on logical negation.} To further understand the FOL reasoning ability of prompt-based methods, we conduct experiments on each form of logic, whose performance and variance of different prompt templates are shown in Figure~\ref{fig2}. We follow the setting of LogicNLI to disentangle diverse logic forms. From Figure~\ref{fig2}, we observe that prompt-based methods achieve stable performance on six kinds of logic (whose medians are around 65\%, respectively) with a relatively low variance among different prompt templates. Sometimes, their performance on existential quantifier can even exceed 70\%. Nevertheless, these methods cannot deal with logical negation well (whose performance's median is approximately 45\%). Meanwhile, the variance of different templates is also very high, proving that prompt-based methods cannot perform consistently on logical negation. In conclusion, it is evident that logical negation mainly hinders prompt-based methods from making correct and reasonable logical reasoning.

\textbf{Logically negative bias leads to the ineffectiveness of the prompt-based methods.} We further investigate why prompt-based methods have trouble handling logical negation by performing the error analysis. From Figure~\ref{figstat}, we find prompt-based methods are prone to misjudging positive labels as negative labels (TF error, around 70\%), and most TF errors are related to logical negation in statements. To validate the observation, we count the proportion of logically negative statements (statements with "not") related to ground truths and predictions on depth-0, shown in Figure~\ref{figtj}. Like the overall dataset, logically negative statements have almost balanced positive and negative labels (51.2\% and 48.8\%). However, prompt-based methods are likely to predict them as "Contradiction" (64.6\%). Meanwhile, predictions on statements without logical negation are biased towards "Entailment" (61.3\%). We define this phenomenon as logically negative bias.

\begin{figure*}[!tb]
\centering
\includegraphics[width=15cm]{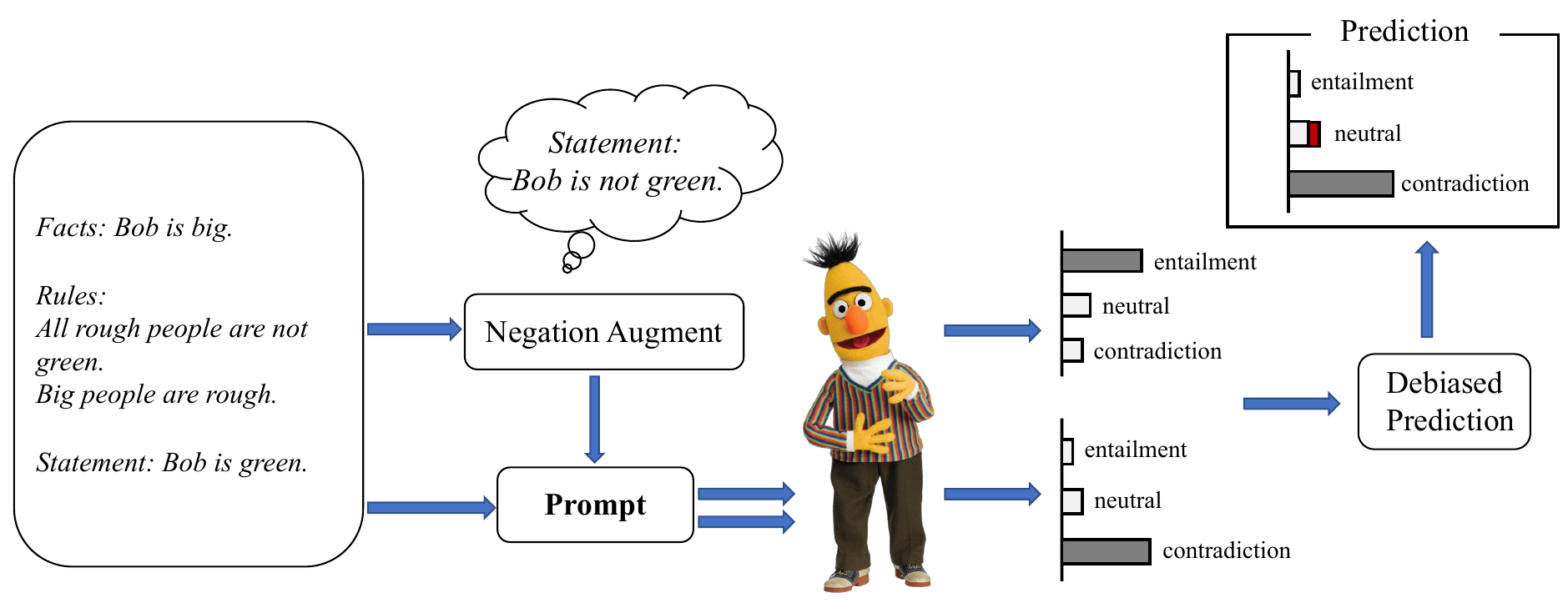}
\caption{Illustration of our method applied to probe a pre-trained language model’s ability to perform FOLs. Each input is placed into the negation augmented and get negative output. The prompts used for both the direct input and the negative output are the same. The red part in the predicted probability graph is the result of debiasing.}
\label{fig3}
\end{figure*}

\section{Negation Augmenting and Negation Debiasing}
\label{sec:all}

To alleviate the impact of logically negative bias, we first make two assumptions to quantify this bias.

\textbf{Assumption 1:} Regarding the prompt-based methods as a probabilistic model, if statements include logical negation, the logically negative bias only increases the probability of the negative label; otherwise, it increases the probability of the positive one. We define such increments introduced by the bias as $\beta_1$ and $\beta_2$.

\textbf{Assumption 2:} The label's probability conditioning on a proposition is equal to the probability of the counterpart label conditional on the negative proposition. $P(y|x) = P(\bar{y}|\neg x)$, where $x$ and $y$ represent the proposition and the label, while $\neg x$ means the negative proposition of $x$ and $\bar{y}$ is the counterpart label of $y$, which can be generated by $y$ and prior knowledge.

Based on these two assumptions, we propose a simple but effective method, negation augmenting and negation debiasing (NAND), to eliminate the logically negative bias of prompt-based methods without model retraining, whose framework is shown in Figure~\ref{fig3}.

\subsection{Negation Augmenting}
According to Assumption 2, negation augmenting (NA) is a simple but effective operation that introduces negative propositions and counterpart labels into prompt-based methods. In NLI settings, $x$ represents an instance with facts, rules, and a statement, while $y$ is the label, where $y \in Y=\{E, C, N\}$, where $E$, $C$, and $N$ represents "Entailment", "Contradiction", and "Neutral" respectively. The definition of $Y$ is shown in Equation~\ref{eq1},  where $\Psi$ means all facts and rules, $s$ means the judgment statement, and $\vdash$ means syntactic consequence.

\begin{equation}
    \small
    y = \left\{\begin{array}{ll} 
            E, & \Psi \rightarrow s \\
            C, & \Psi \rightarrow \neg s \\
            N, & \Psi \nrightarrow s \wedge \Psi \nrightarrow \neg s \\
        \end{array}\right.
    \label{eq1}
\end{equation}

It is intuition that if we wonder whether a statement is "Entailment", we will check whether its negative statement is "Contradiction" under the same facts and rules. Inspired by this intuition, the negation of an instance $\neg x$ can be acquired by designing a logically negative statement and keeping facts\&rules the same. In particular, we turn double negative statements into positive ones, preventing the introduction of more negative biases. As a result, the counterpart label $\bar{y}$ can be acquired through Equation~\ref{eq2} naturally.

\begin{equation}
    \small
    \bar{y}(y) = \left\{\begin{array}{ll} 
            E, & y = C \\
            C, &  y = E \\
            N, & y = N \\
        \end{array}\right.
    \label{eq2}
\end{equation}

For each instance x, the template $T()$ is used to map $x$ to the prompt input $T(x)$. Therefore, the original predicted label $y_p$ can be calculated by Equation~\ref{eq3}, where $P(y_p|T(x))$ maps prompts to a probability distribution with the softmax function.

\begin{equation}
  \small
  y_p = {{\arg\max}_{y \in Y} \, P(y|T(x))}
  \label{eq3}
\end{equation}

NA further takes $\neg x$ and $\bar{y}$ into consideration to make the decision of $y_p$. For the example in Figure~\ref{fig3}, when determining whether "Bob is green" is "Contradiction", NA introduces "Bob is not green" and we should also consider such the negation as "Entailment". Therefore, the calculation of $y_p$ with NA can be rewritten in Equation~\ref{eq4}.

\begin{equation}
\small
   y_p = {{\arg\max}_{y \in Y} \{ \, P(y|T(x))+P(\bar{y}(y)|T(\neg x))\}}
   \label{eq4}
\end{equation}

This union method meets the definition that logical negation is an operation on the true value, which is different from the original one that only manages superficial correlations between negative words and conclusions. Intuitively, NA introduces logical negation to all instances to remove the logically negative bias. We can further qualitatively analyze the effectiveness of NA roughly based on two assumptions.

According to Assumption 1, if statements do not contain "not", the unnormalized probabilities of three labels are $P_E + \beta_1$, $P_C$, and $P_N$; otherwise, unnormalized probabilities are $P_E$, $P_C + \beta_2$, and $P_N$, respectively. the unnormalized probability is the output before the "softmax" layer. Due to the monotonically increasing nature of the "softmax" function, unnormalized probabilities are reasonable in the "argmax" computation. According to Assumption 2, unnormalized probabilities after NA are $2 P_E + \beta_1$, $2 P_C + \beta_2$, and $2 P_N$.

Considering the situation with logical negation, NA aims to revise TF errors, where $P_E > P_C$ and $P_E < P_C + \beta_2$. Therefore, $2 P_E + \beta_1 > 2 P_C + \beta_2$. The simplified formula is shown in Equation~\ref{eq5}. Only if the probabilities satisfy the condition in the equation can TF errors be corrected. Similarly, the constraint to revise FT errors is shown in Equation~\ref{eq6}. From the two equations, we can observe that the more serious the logically negative bias is, and the closer the biased levels are, the better the correction ability of NA is. Meanwhile, NA does not introduce any other errors, so NA always brings non-negative gains to prompt-based methods.

\begin{equation}
\small
  \frac{\beta_2 - \beta_1}{2} < P_E - P_C < \beta_2
  \label{eq5}
\end{equation}

\begin{equation}
\small
  \frac{\beta_1 - \beta_2}{2} < P_C - P_E < \beta_1
  \label{eq6}
\end{equation}

In addition, NA fits OWA because of the definition of $\bar{y}$. Considering to CWA, the world assumption that do not distinguish between the negative and uncertain labels, $\bar{y} = Entailment$ will not always hold even if $y = Contradiction$. As a result, we adopt the degenerate NA that only imposes $\neg x$ on the probability of "Entailment". We show the replacement of degenerate NA in Equation~\ref{cwa}, where $\leftarrow$ means the replacement operation.

\begin{equation}
    \small
    \begin{array}{l} 
            P(y = E|T(x)) \leftarrow P(y = E|T(x)+P(y = C|T(\neg x), \\
            P(y = C|T(x)) \leftarrow 2*P(y = C|T(x)) \\
    \end{array}
    \label{cwa}
\end{equation}

\subsection{Negation Debiasing} 
Although NA dilutes the effect of the logically negative bias, it cannot remove the bias between "Entailment"/"Contradiction" and other labels ("Neutral"). To remedy this deficiency, we propose a negative debiasing method (ND). Specifically, after NA, the unnormalized probabilities of three labels are $2P_E + \beta_1$, $2P_C + \beta_2$, and $2 P_N$, and we are required to introduce an offset $\gamma$ to adjust $2 P_N$ to $2 P_N + \gamma$ effectively. The only remained problem is how to construct $\gamma$ satisfying two conditions: 1) offset $\beta_1$ and $\beta_2$ as much as possible; 2) do not introduce new bias.

The first condition requires ND to correct errors of mispredicting "Neutral" to be "Entailment"/"Contradiction", where $P_N > P_E (P_C)$ and $ 2 P_N < 2 P_E (2 P_C) + \beta_1 (\beta_2)$. Therefore, $2 P_E + \gamma > 2 P_E (2 P_C) + \beta_1 (\beta_2)$ should hold. After simplification, two conditions are shown in Equation~\ref{eq7}. Theoretically, the larger $\gamma$, the more errors ND can correct.

\begin{equation}
    \small
    \left\{\begin{array}{ll} 
            P_N - P_E < \beta_1 < 2 P_N - 2 P_E + \gamma \\
            P_N - P_C < \beta_2 < 2 P_N - 2 P_C + \gamma \\
        \end{array}\right.
    \label{eq7}
\end{equation}

However, considering the second condition, $\gamma$ should be limited. To avoid errors of mispredicting "Entailment"/"Contradiction" to be "Neutral", $P_E (P_C) > P_N$ and $2 P_E (2 P_C) + \beta_1 (\beta_2) > 2 P_N + \gamma$ should hold. Strictly, $\gamma \leq \min\{\beta_1, \beta_2\}$ satisfies the condition. Combining the two conditions, the optimal solution of $\gamma$ is $\min\{\beta_1, \beta_2 \}$.

\subsection{A General, Empirical Estimation of $\gamma$} 
To determine $\gamma$, we should first estimate $\beta_1$ and $\beta_2$. Practically, we use normalized probabilities to estimate two variables, which are equivalence to unnormalized probabilities. We assume that $\beta_1 \sim \mathcal{N}(\mu_1, \sigma_1)$ and $\beta_2 \sim \mathcal{N}(\mu_2, \sigma_2)$, where $\mathcal{N}$ means Gaussian distribution, $\mu$ and $\sigma$ are mean and standard deviation, respectively. According to Assumption 2, $\mu$ and $\sigma$ can be roughly estimated by the difference between $P(y|x)$ and $P(\bar{y}|\neg x)$. After estimation, we adopt the 2-$\sigma$ principle that $\gamma \approx \min\{\mu_1 - 2 \sigma_1, \mu_2 - 2 \sigma_2 \}$. This principle guarantees more than 95\% instances match ND. In reality, such the ratio is higher due to the $\min$ operation, so $\gamma$ hardly introduces new errors. We can also find that the more serious the logically negative bias is, and the smaller the bias variance is, the better the correction ability of ND is.


\begin{table*}[!tb] 
    \centering 
    \fontsize{8}{15}\selectfont    
    \begin{threeparttable} 
        \begin{tabular}{ccccccccc}  
            \toprule         
            \multirow{2}{*}{}& \multirow{2}{*}{\bf Models/ Acc.(\%)}&\multicolumn{7}{c}{\bf RuleTaker(CWA)}\\
            & & depth-0 & depth-1 & depth-2 & depth-3 & depth-3ext & depth-3ext-NatLang & depth-5 \\
            \midrule 
            {\bf Random}& & 50.0 & 50.0 & 50.0 & 50.0 & 50.0 & 50.0 & 50.0 \\
            \cmidrule(lr){1-9}
            \multirow{2}{*}{\bf FT} & BERT & \underline{99.8} & 89.3 & 81.3 & 72.5 & 69.3 & 65.2 & 62.5 \\
            & RoBERTa & \underline{99.8} & 92.1 & 85.2 & 81.6 & 76.1 & 75.4 & 73.0\\
            \cmidrule(lr){1-9}
            \multirow{2}{*}{\bf Prompt}& BERT & 62.9 & 57.3 & 60.2 & 59.8 & 59.8 & 59.5 &  61.5\\
            & RoBERTa & 73.8 & 66.7 & 67.0 & 67.3 & 67.7 & 68.9 & 69.8 \\
            \multirow{2}{*}{\bf AutoPrompt}& BERT & \textbf{66.3} & 55.7 & 53.2 & 52.1 & 53.3 & 52.6 &  51.3\\
            & RoBERTa & \textbf{75.7} & 55.6 & 56.5 & 53.8 & 55.1 & 55.4 & 52.6 \\
            \multirow{2}{*}{\bf +NA(ours)} & BERT & 64.7 & \textbf{62.1} & \textbf{63.2} & \textbf{61.1} & \textbf{63.0} & \textbf{63.7} & \textbf{64.4} \\
            & RoBERTa & 75.4& \textbf{69.0} & \textbf{68.1} & \textbf{68.6} & \textbf{70.4} & \textbf{71.1} & \textbf{72.2} \\
            \bottomrule 
        \end{tabular}
    \end{threeparttable}
    \caption{Results of RuleTaker under CWA. ND is not imposed on this task because CWA does not contain "Neutral" labels. If there is more than one prompts, we show the best one among diverse prompt formats. \textbf{Underlined results represent the in-domain tests that models have supervised trained on the same dataset, while all others are the out-domain results.} The depth-3ext-NatLang dataset is an augmented one in crowdsourced natural language based on depth-3ext.}
    \label{Results1}
\end{table*}

\section{Experiments}

\begin{table*}[h] 
    \centering 
    \fontsize{7}{14}\selectfont    
    \begin{threeparttable} 
        \begin{tabular}{cccccccccc}  
            \toprule         
            \multirow{3}{*}{}& \multirow{3}{*}{\bf Models/ Acc.(\%)}&\multicolumn{8}{c}{\bf Datasets}\\
            \cmidrule(lr){3-10}
            & & \multicolumn{7}{c}{ProofWriter} & LogicNLI \\
            & & depth-0 & depth-1 & depth-2 & depth-3 & depth-3ext & depth-3ext-NatLang & depth-5 & depth$\leq$5 \\
            \midrule 
            {\bf Random}& & 33.3 & 33.3 & 33.3 & 33.3 & 33.3 & 33.3 & 33.3  & 33.3 \\
            \cmidrule(lr){1-10}
            \multirow{2}{*}{\bf FT} & BERT & \underline{94.8} & 81.5 & 76.4 & 71.3 & 69.7 & 68.0 & 67.5 & \underline{78.9} \\
            & RoBERTa & \underline{96.1} & 83.1 & 79.3 & 77.0 & 76.7 & 75.9 & 72.7 & \underline{85.2}\\
            \cmidrule(lr){1-10}
            \multirow{2}{*}{\bf Prompt}& BERT & 52.0 & 53.6 & 54.8 & 54.4 & 54.0 & 53.4 &  54.1 & 52.1\\
            & RoBERTa & 53.5 & 54.0 & 55.6 & 55.9 & 55.6 & 60.9 & 55.7 & 56.5 \\
            \multirow{2}{*}{\bf +NAND ($\gamma=0$)} & BERT & 54.5 & 54.7 & 56.5 & 57.5 & 57.1 & 56.8 & 57.8 & 54.1\\
            & RoBERTa & 70.4 & 71.5 & 72.1 & 72.0 & 71.9 & 76.7 & 70.7 & 61.7\\
            \multirow{2}{*}{\bf +NAND ($\gamma=0.1$)} & BERT & 60.2 & 60.9 & 61.6 & 64.2 & 64.0 & 63.8 & 61.2 & \textbf{61.4}\\
            & RoBERTa & \textbf{78.9} & \textbf{76.6} & \textbf{76.0} & \textbf{75.1} & \textbf{75.7} & \textbf{79.5} & \textbf{72.9} & \textbf{62.4}\\
            \multirow{2}{*}{\bf +NAND ($\gamma=0.2$)} & BERT & \textbf{71.4} & \textbf{67.7} & \textbf{68.5} & \textbf{67.8} & \textbf{68.0} & \textbf{68.5} & \textbf{69.1} & 57.9\\
            & RoBERTa & 64.5 & 67.2 & 68.1 & 68.5 & 68.1 & 73.8 & 67.6 & 61.2\\
            \bottomrule 
        \end{tabular}
    \end{threeparttable}
    \caption{Results of ProofWriter and LogicNLI under OWA. To show the effectiveness of our estimation method, we use different $\gamma$ to ND achieve ND. If there is more than one prompts, we show the best one among diverse prompt formats. \textbf{Underlined results represent the in-domain tests that the hops of test sets are the same as train sets}, while all others are the out-domain results.}
    \label{Results}
\end{table*}

\subsection{Experimental Settings}

We evaluate the effectiveness of NAND on three datasets, RuleTaker~\cite{DBLP:conf/ijcai/ClarkTR20}, ProofWriter~\cite{DBLP:conf/acl/TafjordDC21}, and LogicNLI~\cite{DBLP:conf/emnlp/TianLCX0J21}. RuleTaker is a CWA dataset, while both ProofWriter and LogicNLI are OWA datasets. Each instance of these three datasets contains multiple rules, facts, and a statement to be judged. In this work, we take the conventional prompt-based method (Prompt)~\cite{DBLP:conf/icml/ZhaoWFK021} and AutoPrompt~\cite{DBLP:conf/emnlp/ShinRLWS20} as few-shot baselines. Prompt templates of the former are the same as those used in Section 4, and more are displayed in Appendix~\ref{parameter}. As for AutoPrompt, we choose only 10\% of the depth-0 data to generate automatic prompts, which is categorized as the few-shot baseline for comparison. Meanwhile, the random guess (Random) and the fine-tuning method(FT) are set as the upper and lower bounds. We use the same PLMs, BERT and RoBERTaBERT~\cite{DBLP:conf/naacl/DevlinCLT19} well as RoBERTa~\cite{DBLP:journals/corr/abs-1907-11692}, for all experiments. Hyper-parameters are given in Appendix~\ref{parameter}. Specially, we show results that train on 0-hop instances, and out-of-domain tests on more hops data(other results on generalization are shown in the Appendix~\ref{result-a}).

\subsection{Results} 

\textbf{CWA.} Results of RuleTaker under CWA are shown in Table~\ref{Results1}. We can observe that the prompt-based method (Prompt) and AutoPrompt outperform the random guess but underperform fine-tuning(FT). AutoPrompt performs better on the depth-0 but performs worse on other data than Prompt. This is due to the bias of the auto-generation prompts towards depth-0. Meanwhile, prompt-based methods on RoBERTa achieve better performance than them on BERT. Furthermore, NA can improve the performance of Prompt. Such improvements are significant enough to exceed four points at most. Meanwhile, the performance of Prompt+NA (+NA) on the most complex dataset (depth-5) reaches 72.2\%, which is close to the fine-tuning method.

\textbf{OWA.} Results of ProofWriter and LogicNLI under OWA are shown in Table~\ref{Results}. Considering LogicNLI,  NAND improves the accuracies from 52.1\% to 61.4\% on BERT and from 56.5\% to 62.4\% on RoBERTa. As for ProofWriter, NAND brings more than 10-point and 15-point improvements on all test sets of BERT and RoBERTa, respectively. It is evident that the effectiveness of NAND is confirmed when the identical model and prompt template are provided. In addition, the gap between FT and Prompt+NAND on LogicNLI is still large. However, as the number of hops increases, the performance of Prompt+NAND on ProofWriter gradually approaches FT or even exceeds it (depth-3ext-NatLang and depth-5). This phenomenon also proves that NAND can enhance the generalization ability of Prompt.

Overall, NAND is indeed an effective method to solve logically negative bias and enhance the logical reasoning ability. In fact, NA can alleviate the bias between positive and negative labels so that it can bring improvements under CWA. On the other hand, ND is essential to offset the bias between positive/negative labels and other labels, and it thus brings further improvements under OWA. Although the levels of improvement on diverse conditions are different, NAND hardly brings negative gains for prompt-based methods. In addition, an interesting point is that NAND achieves more benefits on the OWA dataset (ProofWriter) than on the CWA dataset (RuleTaker) with the same setting, meaning that NAND can better deal with situations with uncertainty.

\subsection{Effectiveness Analysis of NAND}
\begin{figure}[t]
\centering
\includegraphics[width=7.5cm]{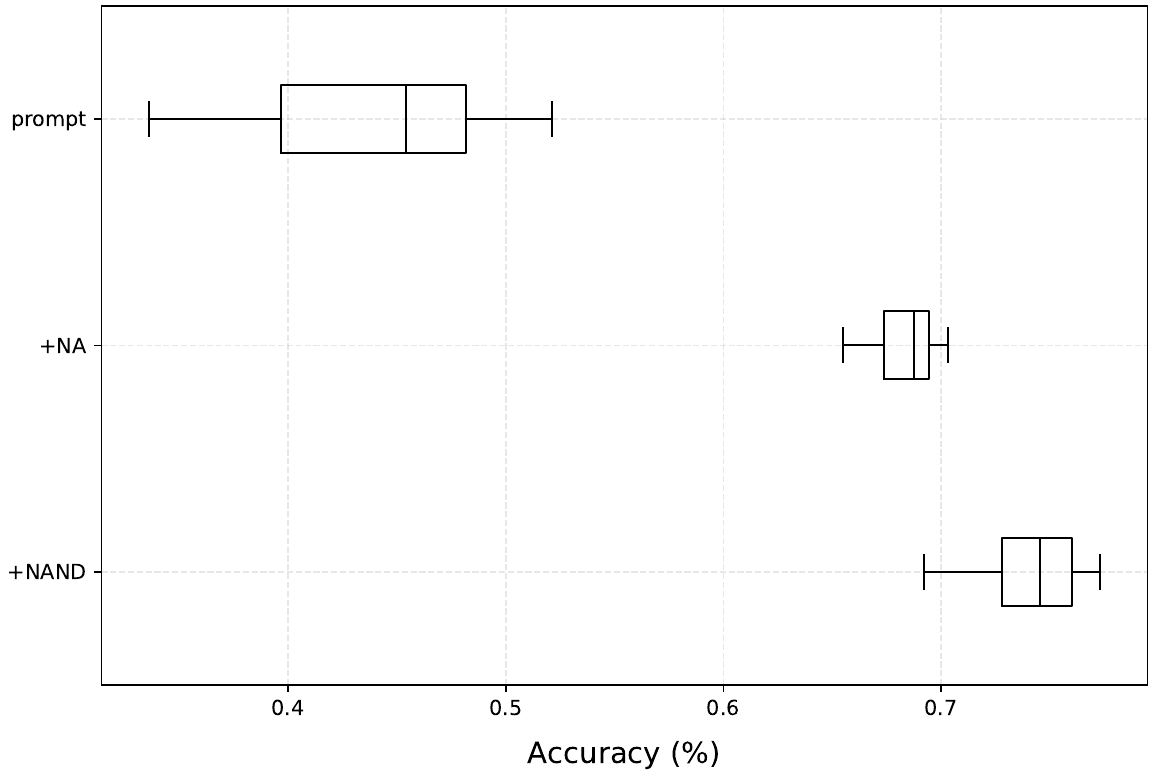}
\caption{Performance on negation logic (\%). We plot the accuracy on negation logic after negation augmenting and debiasing. We use the BERT and the same prompt templates as Figure~\ref{fig2}.}
\label{fig5}
\end{figure}

As we know, a well-designed prompt-based enhancement method should not only improve the performance of specific prompts but also enhance the consistency of different templates. In this part, we further analyze whether NAND can effectively reduce the variance among different templates with logical negation. We use the same experimental setting as the analysis of Figure~\ref{fig2}, while Figure~\ref{fig5} shows the performance distribution of the baselined prompt-based method (Prompt), the NA-enhanced method (+NA), and the NAND-enhanced method (+NAND). From the figure, both NA and NAND significantly enhance Prompt's consistency, which means that different prompt templates result in similar results. Moreover, comparing Prompt+NA and Prompt+NAND, the former shows lower variance, while the latter achieves better performance than the former. In conclusion, NAND not only matches specific templates but is a general method for all effective prompt templates.

\subsection{Analysis on $\gamma$}
According to the estimated methods of $\beta_1$ and $\beta_2$ in Section 5.3, we list the estimation results in Table~\ref{guassian}. Based on the 2-$\sigma$ principle, $\gamma^*_{BERT} \approx \min\{0.276 - 0.053 \times 2, 0.434 - 0.083 \times 2\} = 0.170$ and $\gamma^*_{RoBERTa} \approx \min\{0.112 - 0.025 \times 2, 0.321 - 0.080 \times 2\} = 0.062$. From Table~\ref{Results}, we can observe the closer the $\gamma$ is to the estimated value, the better the performance of the NAND. 

In practice, we set $\gamma > \gamma^*$, which cannot ensure that $\gamma < \beta_1$ and $\gamma < \beta_2$ for most instances. However, ND still works. This is because the 2-$\sigma$ is a strict, general principle to guarantee the lower bound of ND. Sometimes, the 1-$\sigma$ or other principles may result in better performance but cannot be generalized to other situations. Nevertheless, if $\gamma > \min\{\mu_{\beta_1}, \mu_{\beta_2}\}$, ND will inevitably introduce other biases that make its effects uncontrollable, like $\gamma_{RoBERTa} = 0.2$. 

\begin{table}[!tb] 
    \centering 
    \fontsize{10}{16}\selectfont    
    \begin{threeparttable} 
        \begin{tabular}{cccc}  
            \toprule     
            Bias Estimation & PLM & $\mu$ & $\sigma$ \\
            \midrule 
            \multirow{2}{*}{$\beta_1$} & BERT & 0.276 & 0.053 \\
            & RoBERTa & 0.112 & 0.025\\
            \cmidrule(lr){1-4}
            \multirow{2}{*}{$\beta_2$} & BERT & 0.434 & 0.083\\
            & RoBERTa & 0.321 & 0.080\\
            \bottomrule 
        \end{tabular}
    \end{threeparttable}
    \caption{Bias estimations of $\beta_1$ and $\beta_2$. }
   \label{guassian}
\end{table}

\section{Conclusion}
In this paper, we study the effectiveness of the prompt-based methods on first-order logical reasoning. Through a detailed analysis, we find that the bottleneck lies in logical negation among seven FOLs and arises from logical negation bias. To solve the problem, we propose a simple but effective method, Negation Augmenting and Negation Debiasing (NAND). Experiments show that NAND can improve the logical negation ability and help prompt-based methods of logical reasoning.

\section*{Limitations}
During our experiments and analysis, the influence of manually prompted cases could not be completely ruled out. Although we use a variety of commonly used prompt templates, we cannot guarantee that we will find the best hint. However, we try to reduce the gap with auto-generated prompts by AutoPrompt as much as possible. Meanwhile, incorporating logical negation operations into prompt learning does not work well for CWA. CWA is an assumption that is not commonly used and it does not distinguish between negation and uncertainty. After negating on "Contradiction" examples, we can not provide the corresponding pairing labels. Therefore, we use an alternative method.

\bibliography{anthology,custom}
\bibliographystyle{acl_natbib}

\appendix

\section{Training Detail}
\label{parameter}
All of our experiments were conducted using a single GPU with 24GB RAM (NVIDIA GeForce GTX 3090 Ti), including the in-domain pretraining experiment.
\subsection{Versions of major packages}
python: 3.7 \\
pytorch: 1.4\\
transformers: 3.0\\
\subsection{Hyper-parameters}
We have trained each model more than three times. The hyper-parameters of AutoPrompt are the same as the original paper and the hyper-parameters about LMs are as follows:
\begin{table}[!h]
\setlength{\belowcaptionskip}{-0.5cm}
\centering
\begin{tabular}{ccc}
\toprule
\textbf{Paras.} & \textbf{BERT} & \textbf{RoBERTa}\\
\midrule
batch size & $16$ & $16$ \\
lr & $ 5e^{-6} $ & $5e^{-6}$ \\
decay rate & $0.9$ & $0.9$ \\
l2 coeff. & $1e^{-5}$ & $1e^{-5}$ \\
early stop & $7$ & $7$ \\
epochs & $20$ & $20$ \\
optimizer & ADAMW & ADAMW \\
\bottomrule
\end{tabular}
\caption{Hyper-parameter settings.}
\label{tab:Tabel 1}
\end{table}

\subsection{Prompt Examples}
Here, we provide some examples: 1) [CLS] Facts, Rules [SEP] It was [MASK]. Statement. 2) [CLS] Facts, Rules? [SEP] [MASK] Statement. 3) [CLS] Facts, Rules? [SEP] It is [MASK] that Statement. 4) [CLS] Facts, Rules? [MASK], Statement. 5) ...

\section{Datasets}
\begin{table}[h]
\setlength{\belowcaptionskip}{-0.5cm}
\centering
\small
\begin{tabular}{cc}
\toprule
\textbf{Datasets} & \textbf{Link} \\
\midrule
RuleTaker & https://allenai.org/data/ruletaker \\
ProofWriter & https://allenai.org/data/proofwriter  \\
LogicNLI & https://github.com/omnilabNLP/LogicNLI \\
\bottomrule
\end{tabular}
\caption{Download links for all datasets}
\label{tab:data}
\end{table}
\label{dataset}
An overview of the links to downloadable versions of all used datasets can be found in Table~\ref{tab:data}. \textbf{RuleTaker}~\cite{DBLP:conf/ijcai/ClarkTR20} contains seven datasets under a closed-world assumption. Five of them constrain by the maximum depth of inference required to prove the facts used in its questions (up to depths $D=0$, $D\leq1$, $D\leq2$, $D\leq3$ and $D\leq5$ respectively).
Another two are hand-authored reasoning datasets independently to further test the robustness and out-of-distribution performance. RuleTaker concentrates on a specific FOL combination, conjunctive implication with negation. \textbf{ProofWriter}~\cite{DBLP:conf/acl/TafjordDC21} is based on the RuleTaker research line and includes seven datasets. However, it makes an open-world assumption that distinguishes between negation and uncertainty.  \textbf{LogicNLI}~\cite{DBLP:conf/emnlp/TianLCX0J21} is a logically more sophisticated dataset with a large number of logical expressions that span all seven FOLs under OWA. Besides, LogicNLI includes separate test sets for seven types of logic (only implication and equivalence can be fully entangled from other FOLs and directly used for reasoning, while others alone cannot constitute complete reasoning).

\section{Generalization Experiments}
\begin{figure}[t]
\centering
\subfigure[Results on RuleTaker]{
\begin{minipage}[b]{0.5\textwidth}
\includegraphics[width=1\textwidth]{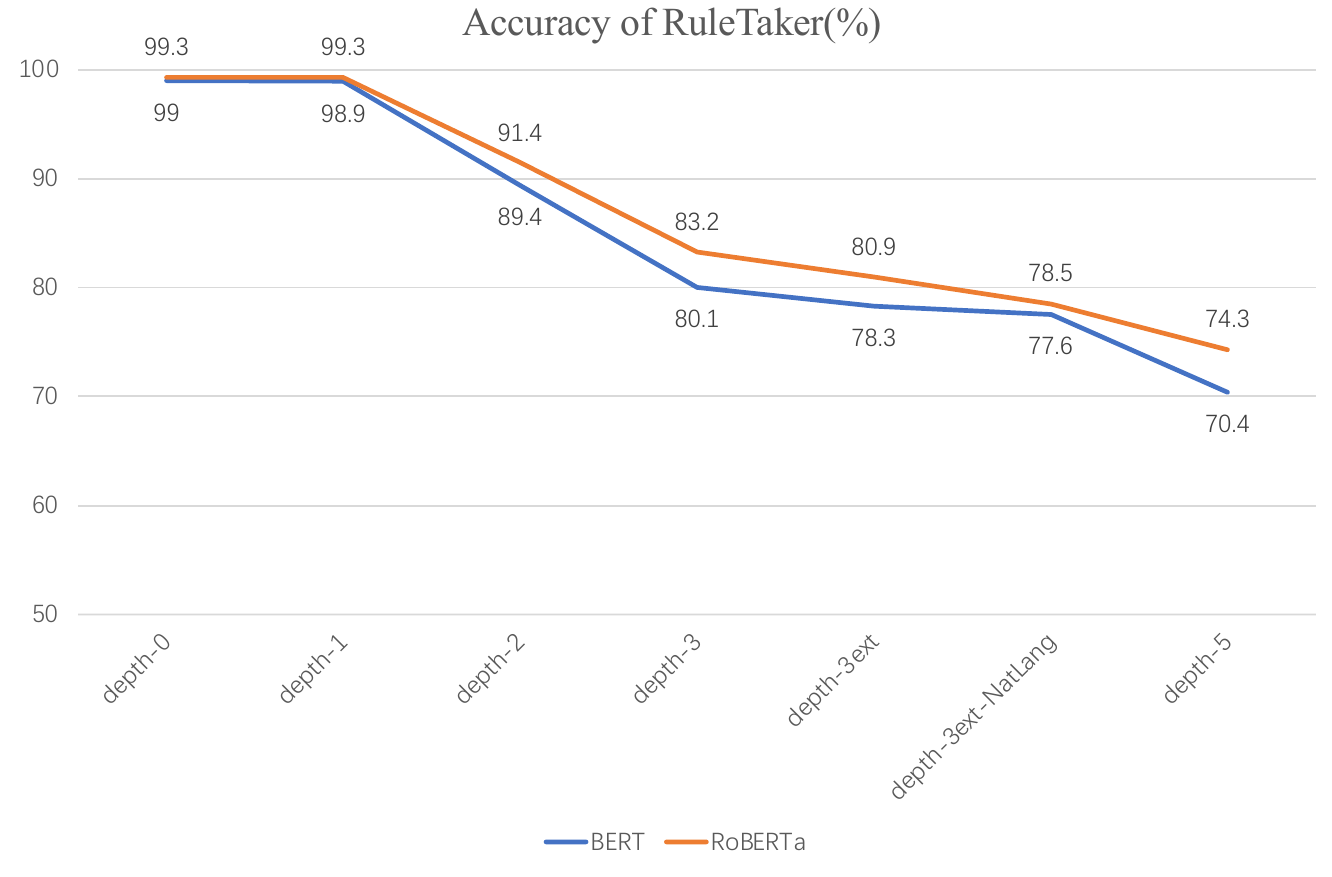}
\end{minipage}}
\subfigure[Results on ProofWriter ]{
\begin{minipage}[b]{0.5\textwidth}
\includegraphics[width=1\textwidth]{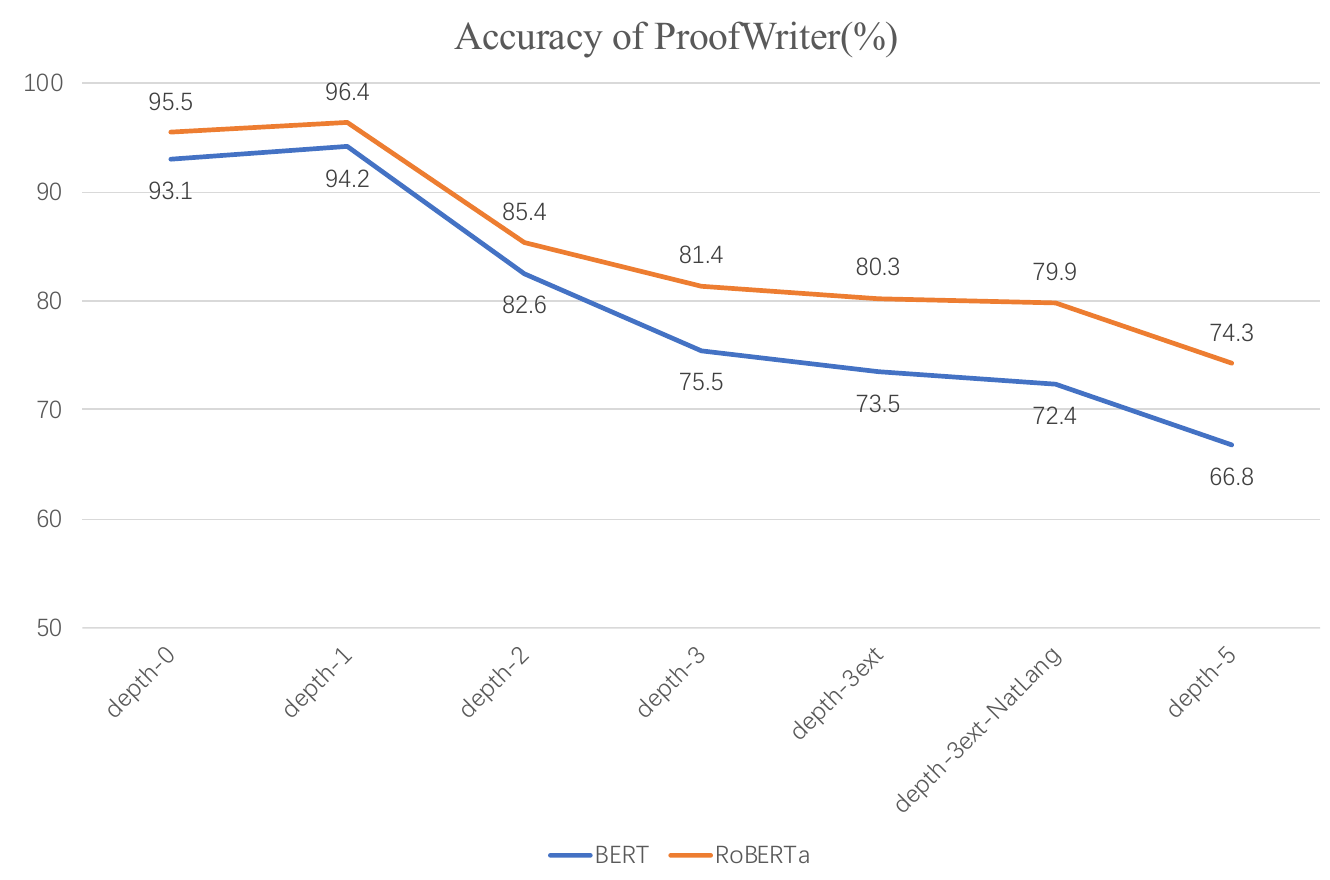}
\end{minipage}}
\caption{Generalization results on RuleTaker and ProofWriter. We further train on the one-hop datasets and test the accuracy variation of the models.} 
\label{case_2}
\end{figure} 
\label{result-a}
We further train on the one-hop datasets and test the accuracy variation of both BERT and RoBERTa. As results on Figure~\ref{case_2}(a) and Figure~\ref{case_2}(b) shown, both BERT and RoBERTa performance drops significantly as the number of inference hops increases.

\end{document}